\documentclass[letterpaper]{article}
\usepackage{aaai}
\usepackage{times}
\usepackage{helvet}
\usepackage{courier}
\usepackage{amsmath}
\usepackage{graphicx}
\usepackage[linesnumbered]{algorithm2e}
\usepackage{url}
\usepackage{tabularx}
\usepackage[table]{xcolor}
\usepackage{bm}
\usepackage{amssymb}
\begin{document}
%
\title{Methods for Integrating Knowledge with the Three-Weight Optimization Algorithm for Hybrid Cognitive Processing}
\author{Nate Derbinsky \and Jos\'{e} Bento \and Jonathan S. Yedidia\\
Disney Research\\
222 Third Street\\
Cambridge, MA 02142\\
\{nate.derbinsky, jbento, yedidia\}@disneyresearch.com
}
\maketitle
\begin{abstract}
\begin{quote}
In this paper we consider optimization as an approach for quickly and flexibly developing hybrid cognitive capabilities that are efficient, scalable, and can exploit knowledge to improve solution speed and quality.
In this context, we focus on the Three-Weight Algorithm, which aims to solve general optimization problems.
We propose novel methods by which to integrate knowledge with this algorithm to improve expressiveness, efficiency, and scaling, and demonstrate these techniques on two example problems (Sudoku and circle packing).
\end{quote}
\end{abstract}


\noindent A central goal of cognitive-architecture research is to integrate in a task-independent fashion the broad range of cognitive capabilities required for human-level intelligence, and a core challenge is to implement and interface the diverse processing mechanisms needed to support these capabilities.

The Soar cognitive architecture \cite{Laird2012} exemplifies a common approach to this problem: Soar integrates a \emph{hybrid} set of highly \emph{specialized} algorithms, which leads to \emph{flexibility} in the types of knowledge about which it can reason and learn; \emph{efficiency} for real-time domains; and \emph{scalability} for long-lived agents in complex environments.
However, because each algorithm is highly optimized, it can be challenging to experiment with architectural variants.

By contrast, work on the Sigma ($\Sigma$) architecture \cite{Rosenbloom2011Rethinking} has exemplified how hybrid cognitive capabilities can arise from \emph{uniform} computation over tightly integrated graphical models.
When compared with Soar's hybrid ecosystem, this approach allows for comparable flexibility but much improved speed of integrating and experimenting with diverse capabilities.
However, utilizing graphical models as a primary architectural substrate complicates the use of rich knowledge representations (e.g. rules, episodes, images), as well as maintaining real-time reactivity over long agent lifetimes in complex domains \cite{Rosenbloom2012}.

This paper takes a step towards an intermediate approach, which embraces a \emph{hybrid} architectural substrate (ala Soar), but seeks to leverage \emph{optimization} over factor graphs (similar to Sigma) via the \emph{Three-Weight Algorithm} (TWA; \citeauthor{Derbinsky2013AnImp} \citeyear{Derbinsky2013AnImp}) as a general platform upon which to rapidly and flexibly develop diverse cognitive-processing modules.
We begin by describing why optimization is a promising formulation for specialized cognitive processing.
Then we describe the TWA, focusing on its generality, efficiency, and scalability.
Finally, we present novel methods for integrating high-level knowledge with the TWA to improve expressiveness, efficiency, and scaling and demonstrate the efficacy of these techniques in two tasks (Sudoku and circle packing).

This paper does \emph{not} propose a new cognitive architecture, nor does the work result from integrating the TWA with an existing architecture.
Instead, we propose a paradigm and a set of methods to enable researchers interested in pursuing integrated cognition.


\section{Optimization}

A general \emph{optimization problem} takes the form

\begin{equation}
\label{eq:opt_problem}
\begin{split}
& \underset{ \bm{v} \in \mathbb{R}^{n} }{ \text{minimize} } : f(\bm{v}) = f_1(v_1, v_2, \dots) + f_2(\dots) + \dots \\
& + \sum{c_k(v_1, v_2, \dots) = { \begin{cases} 0 & \text{constraint met}\\ \infty & \text{else} \end{cases}}}
\end{split}
\end{equation}

\noindent where $f(\bm{v}) : \mathbb{R}^{n} \rightarrow \mathbb{R}$ is the \emph{objective function} to be minimized over a set of variables $\bm{v}$ and $c_k(\dots)$ are a set of \emph{hard constraints} on the values of the variables. By convention we consider \emph{minimization} problems, but \emph{maximization} can be achieved by inverting the sign of the objective function.

As we will exemplify with our discussion of the Three-Weight Algorithm, it is often useful to consider families or classes of optimization problems, which are characterized by particular forms of the objective and constraint functions.
For example, much recent work has been done on \emph{convex} optimization problems, in which both the objective and constraint functions are convex \cite{Boyd2004Convex}.
However, neither the TWA nor our proposed approach are constrained to any class of optimization problem.

\begin{figure}[t]
\begin{center}
\includegraphics[width=.6\columnwidth]{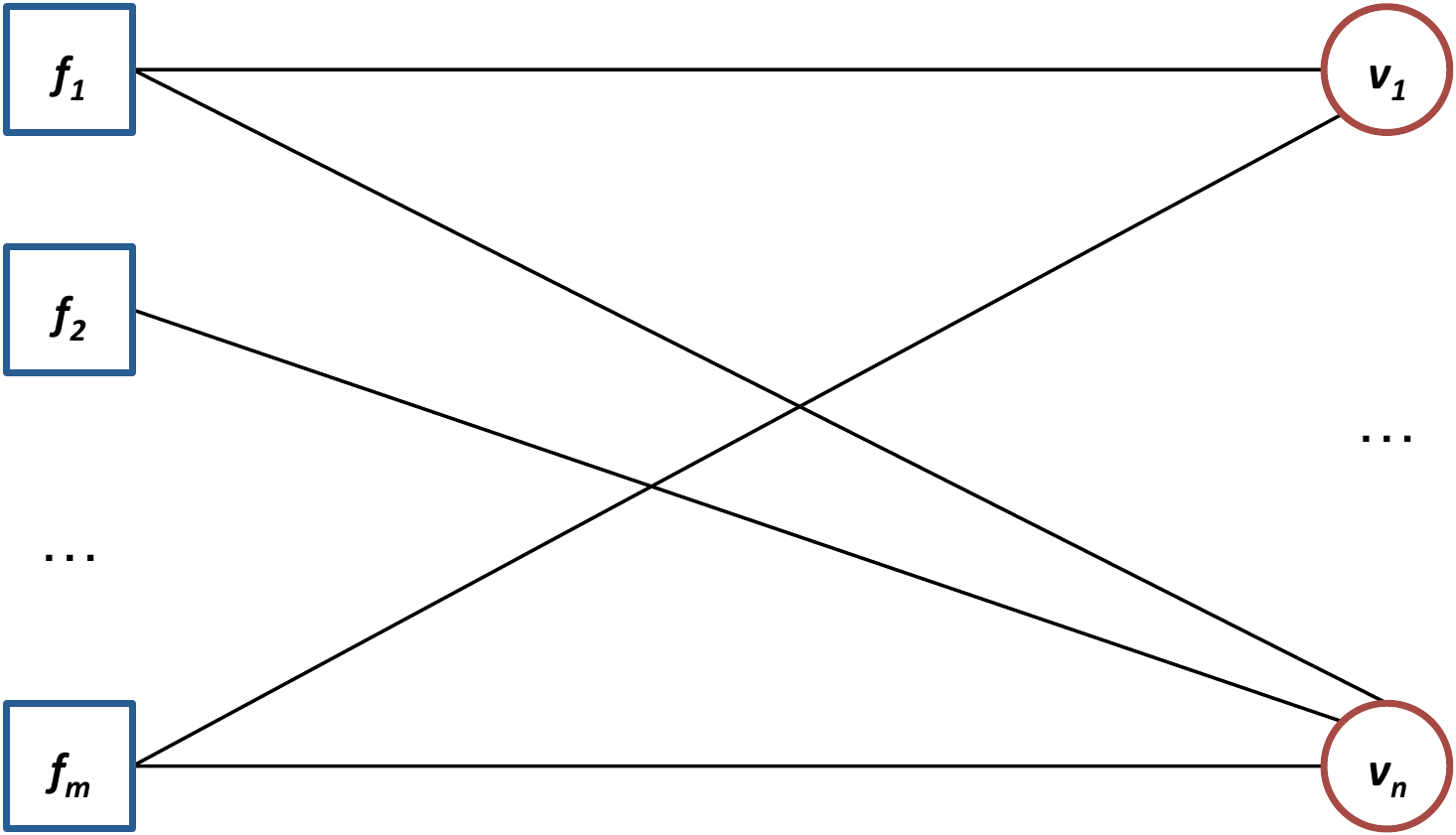}
\end{center}
\caption{Factor graph of an optimization problem whose objective function is $f(\textbf{v})=f_1(v_1, v_n)+f_2(v_n)+\ldots+f_m(v_1, v_n)$. The red circles represent the variables, while the blue squares represent hard or soft cost functions. If a line connects a square to circle, that means that the corresponding cost function depends on the corresponding variable.}
\label{fig:factorgraph}
\end{figure}

Optimization is a useful framework in the context of hybrid cognitive processing for two primary reasons: (1) generality of problem formulation and (2) independence of objective function and solution method.
First, the form in Equation \ref{eq:opt_problem} is fully general, supporting such diverse processing as constraint satisfaction (a problem with only hard constraints, as we illustrate with Sudoku) and vision/perception \cite{Geman1984Stochastic}.
Often these problems are represented as a \emph{factor graph} \cite{Kschischang2001}, as exemplified in Figure \ref{fig:factorgraph}.
Like other graphical models, factor graphs decompose the objective function into independent hard or soft cost functions, thereby reducing the combinatorics that arise with functions of multiple variables.

Another important reason to embrace an optimization framework is that the objective function is formulated independently from the method by which the the corresponding problem is solved.
This abstraction supports flexibility in experimenting with objective variants without requiring significant effort to change a corresponding algorithm.
However, objective-function changes may impact the \emph{speed} and \emph{success rate} of a particular optimization algorithm, and thus it is advantageous to use an optimization algorithm that can specialize to particular classes of objective functions, as well as adapt solving strategies when provided higher-level sources of knowledge (issues we discuss in greater depth later).

\subsubsection{Related Work}
Broadly speaking, optimization has been applied in three main ways within the cognitive-architecture community.
First, optimization has been applied as a methodological framework with which to rigorously traverse large modeling-parameter spaces, such as the spaces of reward signals \cite{Singh2010Intrinsically} and behavioral strategies \cite{Howes2007Bounding}.
When applied in this way, the particulars of the optimization algorithm, as well as integration within an agent architecture, are typically unimportant, whereas these issues are of primary focus in this paper.
A second, related application of optimization has been as an analytical framework of behavior, as best exemplified by such theories as \emph{rational analysis} \cite{Anderson1991Arch}, \emph{bounded rationality} \cite{Simon1991}, and \emph{optimality theory} \cite{Smolensky2011}.
But again, this utilization of optimization typically does not necessitate a particular implementation-level algorithm, but instead offers evidence and organization for a set of functional capabilities.
Finally, work on Sigma \cite{Rosenbloom2011Rethinking} formulates the entire agent/architecture as an optimization/inference problem.
This paper is much more narrowly scoped in comparison: we discuss optimization as an enabling platform for one or more cognitive modules, independent of the implementation commitments of the architecture as a whole.
However, a sub-problem when considering architecture as an optimization problem is how to formulate individual modules, and thus we revisit this comparison in the next section when discussing the specifics of the Three-Weight Algorithm.


\section{The Three-Weight Algorithm (TWA)}
The Three-Weight Algorithm \cite{Derbinsky2013AnImp} is based on a message-passing interpretation of the Alternating Direction Method of Multipliers, or ADMM, an algorithm that has gained much attention of late within the convex-optimization community as it is well-suited for distributed implementations \cite{Boyd2011}.
The TWA exhibits several properties that make it ideal for cognitive systems:

\begin{itemize}

\item{\textbf{General}. The TWA operates on arbitrary objective functions (e.g. non-linear, non-convex), constraints (hard, soft, mixed), and variables (discrete, continuous). It is known to converge to the global minimum for convex problems and in general, if it converges, the TWA will have arrived at a feasible solution (i.e. all hard constraints met).}

\item{\textbf{Interruptible}. The TWA is an iterative algorithm and, for many problems, intermediate results can serve as heuristic input for warm-starting complementary approaches.}

\item{\textbf{Scalable and Parallelizable}. The TWA takes the form of a \emph{decomposition-coordination} procedure, in which the solutions to small local subproblems are coordinated to find a solution to a large global problem. \citeauthor{Boyd2011} \shortcite{Boyd2011} showed that this algorithmic structure leads naturally to concurrent processing at multiple levels of execution (e.g. MapReduce, multi-core, GPU).}

\end{itemize}

\citeauthor{Derbinsky2013AnImp} \shortcite{Derbinsky2013AnImp} provide a full TWA description, as well as its relationship to ADMM; however, the core of the algorithm (see Algorithm \ref{alg:twa}) can be interpreted as an iteration loop that consists of two phases: (1) \emph{minimizing} each cost function locally and then (2) \emph{concurring} on the result of the local computations.
Importantly, TWA messages operate on the \emph{edges} of the corresponding factor graph, as opposed to directly on variables: this distinction raises the dimensionality of the problem, allowing the TWA to more effectively search the variable space and avoid invalid solutions.

The minimization phase (line 4 in Algorithm \ref{alg:twa}) takes as input a  $({msg}, {weight})$ pair for each edge and must produce, for each edge, a corresponding output pair.
The minimization routine must select a variable assignment in

\begin{equation}
\underset{\bm{v}} {\rm{argmin}} \, \left[ f(\bm{v}) + \frac{{\bm{weight_{in}}}}{2}(\bm{v}-\bm{msg_{in}})^2 \right]
\end{equation}

\noindent The set of variable values must jointly minimize the sum of the local cost function while remaining close to the incoming message set, as balanced by each edge's weight.
Furthermore, each edge must be assigned an outgoing weight, which can be either $0$ (intuitively \emph{no opinion} or \emph{uncertain}), $\infty$ (\emph{certain}), or a ``standard'' weight (typically 1.0).
Proper use of these weight classes can lead to dramatic performance gains \cite{Derbinsky2013AnImp} and is crucial for integration with higher-level knowledge (as discussed later).
The logic that implements this minimization step may itself be a general optimization algorithm, but can also be customized to each cost function; custom minimization almost always leads to improvements in algorithm performance and is typically the bulk of the implementation effort.

The \emph{concur} phase (line 10) combines incoming messages about each variable from all associated local cost functions and computes a \emph{single} assignment value using a fixed logic routine (typically a weighted average).
After each variable node has \emph{concurred}, it is possible to extract this set of values as the present solution state.

We do not go into the details of computing ``messages'' (lines 6 and 12), but two ideas are crucial.
First, each message incorporates both an assignment value and an accumulation over previous \emph{errors} between the value computed by a \emph{local} cost function (line 4) and the \emph{concurred} variable value (line 10).
Second, due to this error term, each edge, even those connected to the same variable node, often communicates a different message: intuitively, each edge has a different view of the variable as informed by an aggregation over local iteration history.
The TWA has converged (line 15) when outgoing messages from \emph{all} variable nodes do not change significantly between two subsequent iterations.

\begin{algorithm}
\KwIn{problem factor graph}
InitializeMsgs()\;
\While{convergence criteria not met}{
\ForEach{factor}{
ComputeLocalVariableAssignments()\;
\ForEach{edge}{
SendOutgoingMsgAndWeight()\;
}
}
\ForEach{variable}{
Concur()\;
\ForEach{edge}{
SendOutgoingMsgAndWeight()\;
}
}
CheckForConvergence()\;
}
\caption{An abstracted version of the Three-Weight Algorithm for general distributed optimization.}
\label{alg:twa}
\end{algorithm}

All message-passing data is local to each edge within the factor graph, and thus it is trivial to coarsely parallelize the two main phases of the algorithm (i.e. all factors can be minimized in parallel and then all variables can be concurred upon in parallel).
For complex cost functions, fine-grained parallelization within the minimization routine may lead to additional performance gains.

The Sigma architecture uses a related algorithm for inference, the sum-product variant of Belief Propagation (BP; \citeauthor{Pearl1982} \citeyear{Pearl1982}).
Both the TWA and BP are message-passing algorithms with strong guarantees for a class of problems (the TWA is exact for convex problems, BP for cycle-free graphical models) and both have been shown to be useful more broadly.
However, the TWA and BP differ along four important dimensions: (1) TWA messages/beliefs are single values (i.e. minimum-energy state) versus distributions in BP; (2) continuous variables are native to the TWA, whereas special treatment (e.g. discretization) is required in BP, which can lead to scaling limitations; (3) because minimization routines can implement arbitrary logic and must only output messages/weights for each edge, the TWA can handle constraints that would be very difficult to capture in BP; and (4) whereas BP can converge to uninformative fixed points and other invalid solutions, the TWA only converges to valid solutions (though they may be local minima).\footnote{\citeauthor{Derbinsky2013AnImp} \shortcite{Derbinsky2013AnImp} more comprehensively compares ADMM, the TWA, and BP.}


\section{Integrating Knowledge with the TWA}

This section discusses two novel methods to integrate higher-level knowledge into the operation of the Three-Weight Algorithm.
These techniques are general and, when specialized for a particular problem, can lead to improved algorithm efficiency (iterations and wall-clock time), scaling, and expressiveness of constraints and heuristics.

\subsection{Reasoner Hierarchy}

We begin by augmenting the TWA iteration loop in order to introduce a two-level hierarchy of \emph{local} and \emph{global} reasoners, as shown in Algorithm \ref{alg:aug}.
Local reasoners are implemented as a special class of factor within the problem graph.
They are able to send/receive messages like other factors, but incoming message values always reflect the concurred upon variable value.
Their default operation is to send zero-weight messages (i.e. have no impact on the problem), but they can also affect change through non-zero-weight messages.
Furthermore, local reasoners have a \emph{Reason} method, which supports arbitrary logic.
We term this class of reasoner ``local'' because, like other factors, it has a local view of the problem via connected variables (and thus can be executed concurrently); however, the added reasoning step affords communication with \emph{global} reasoners.

Global reasoners are implemented as code modules via a single \emph{Reason} method and are not connected to the problem graph, but instead have a ``global'' view of the problem via access to the concurred values of any variable, as well as any information transmitted via local reasoners.
Global reasoners can affect the problem via three main methods: (1) requesting that local reasoners send non-zero-weighted messages; (2) detecting problem-specific termination conditions and halting iteration; and (3) modifying the problem graph, as discussed in the next section.

\begin{algorithm}
\KwIn{problem factor graph}
InitializeMsgs()\;
\While{convergence criteria not met}{
\ForEach{factor}{
ComputeLocalVariableAssignments()\;
\ForEach{edge}{
SendOutgoingMsgAndWeight()\;
}
}
\ForEach{variable}{
Concur()\;
\ForEach{edge}{
SendOutgoingMsgAndWeight()\;
}
}
\ForEach{local reasoner}{
Reason()\;
}
\ForEach{global reasoner}{
Reason()\;
}
CheckForConvergence()\;
}
\caption{Extending the TWA with a two-level hierarchy of \emph{local} and \emph{global} reasoners.}
\label{alg:aug}
\end{algorithm}

As alluded to already, a major reason for a hierarchy of reasoners is to exploit parallelism in order to better scale to large problems.
Thus, where possible, local reasoners serve as a filtering step such that global reasoners need not inspect/operate on the full variable set.
In the Sudoku task, for instance, this hierarchy yields an event-based discrimination network, similar to Rete \cite{Forgy1982}, whereby the local reasoners pass along information about \emph{changes} to possible cell states and a global reasoner implements relational logic that would be difficult and inefficient to represent within the [first-order] problem factor graph.

The TWA weights each message to express reliability, and in the context of symbolic systems, it is often useful to extract \emph{certain} information (${weight}=\infty$).
Local reasoners can filter for \emph{certain} information, propagate implications to global reasoning, and express certain results in outgoing messages.
In Sudoku, for example, propagating certainty via the reasoner hierarchy maintains real-time reactivity for very large problem instances by pruning unnecessary constraints.

So, to summarize, the two-level reasoner hierarchy improves the TWA along the following dimensions:

\begin{itemize}
\item{\textbf{Integration}. Global reasoners can implement arbitrary logic, including complex indexing structures, search algorithms, etc. Information is extracted from the problem, optionally filtered through local reasoners; processed via the global reasoner; and then re-integrated through the API of $({msg}, {weight})$ pairs in local reasoners. Certainty weighting allows for fluid integration with processing mechanisms that operate over symbolic representations.}
\item{\textbf{Expressiveness}. The two-level hierarchy supports relational reasoning over the inherently propositional factor-graph representation without incurring combinatorial explosion. The global reasoner can incorporate richer representations, such as rules, higher-order logics, explicit notions of uncertainty, perceptual primitives, etc.}
\item{\textbf{Efficiency \& Scalability}. Operations of the local reasoners are parallelized, just as factor minimization and variable-value concurrence in the problem graph. Furthermore, effective use of local filtering can greatly reduce the set of variables considered by global reasoners, thereby supporting scaling to large, complex problems.}
\end{itemize}

\subsection{Graph Dynamics}

Our second method for integrating knowledge is to support four classes of graph dynamics by global reasoners: (1) adding/removing edges\footnote{Variable nodes are automatically pruned if there are no remaining incoming edges.}, (2) adding/removing factor nodes\footnote{Removing a factor node has the effect of removing any incoming edges.}, (3) adding new variables, and (4) re-parameterizing factors.
These actions allow for adapting the representation of a single problem instance over time [given experience/knowledge], as well as reusing a single mechanism for multiple problem instances.

Removing graph edges and factors has two main effects: (a) [potentially] changing variable assignments and (b) improving runtime performance.
First, if an edge is disconnected from a factor/variable node, the outgoing variable assignments are now no longer dependent upon that input, and therefore the objective cost may yield a different outcome.
Second, while removing an edge is analogous to sending a zero-weight message, the system need no longer expend computation time, and thus wall-clock time, per iteration, may improve, as we see in both evaluation tasks.

The ability to add and remove edges allows the graph to represent and reason about dynamically sized sets of variable.
For example, in the Sudoku task, the TWA considers a set of possible symbols for each cell, and can reduce its option set over time as logically certain assignments are made within the row/column/square.

Supporting factor re-parameterization supports numerous capabilities.
First, it is possible to reflect incremental environmental changes without having to reconstruct the graph, an important characteristic for online systems.
It is also possible to reflect changes to the objective function, which may come about due to environmental change, task transitions, or dynamic agent preferences/goals.
Additionally, re-purposing existing factors helps keep memory costs stable, which supports scaling to large, complex problems.

So, to summarize, graph dynamics improves the TWA along the following dimensions:

\begin{itemize}
\item{\textbf{Integration}. Global reasoners can dynamically re-structure and re-configure the problem graph to reflect changes in the state of the environment, task structure, as well as agent preferences, goals, and knowledge.}
\item{\textbf{Expressiveness}. Changing edge connectivity supports dynamic sets of variables without the necessity of enumerating all possibilities.}
\item{\textbf{Efficiency \& Scalability}. Performance of the TWA iteration loop depends upon the size of the graph, which can be dynamically maintained such as to represent only those factor nodes, variables, and edges that are necessary.}
\end{itemize}


\section{Demonstrations}
We now evaluate the TWA with our novel knowledge-integration techniques in two tasks: Sudoku and circle packing.
These tasks are not intended to represent cognitive processing modules - we are certainly not suggesting that cognitive architectures should have dedicated puzzle-solving capacities!
Rather, these tasks allow us to demonstrate high-level knowledge integration in the TWA, as well as show benefits for expressiveness, efficiency, and scaling.

\begin{figure}[tb]
\begin{center}
\includegraphics[width=.6\columnwidth]{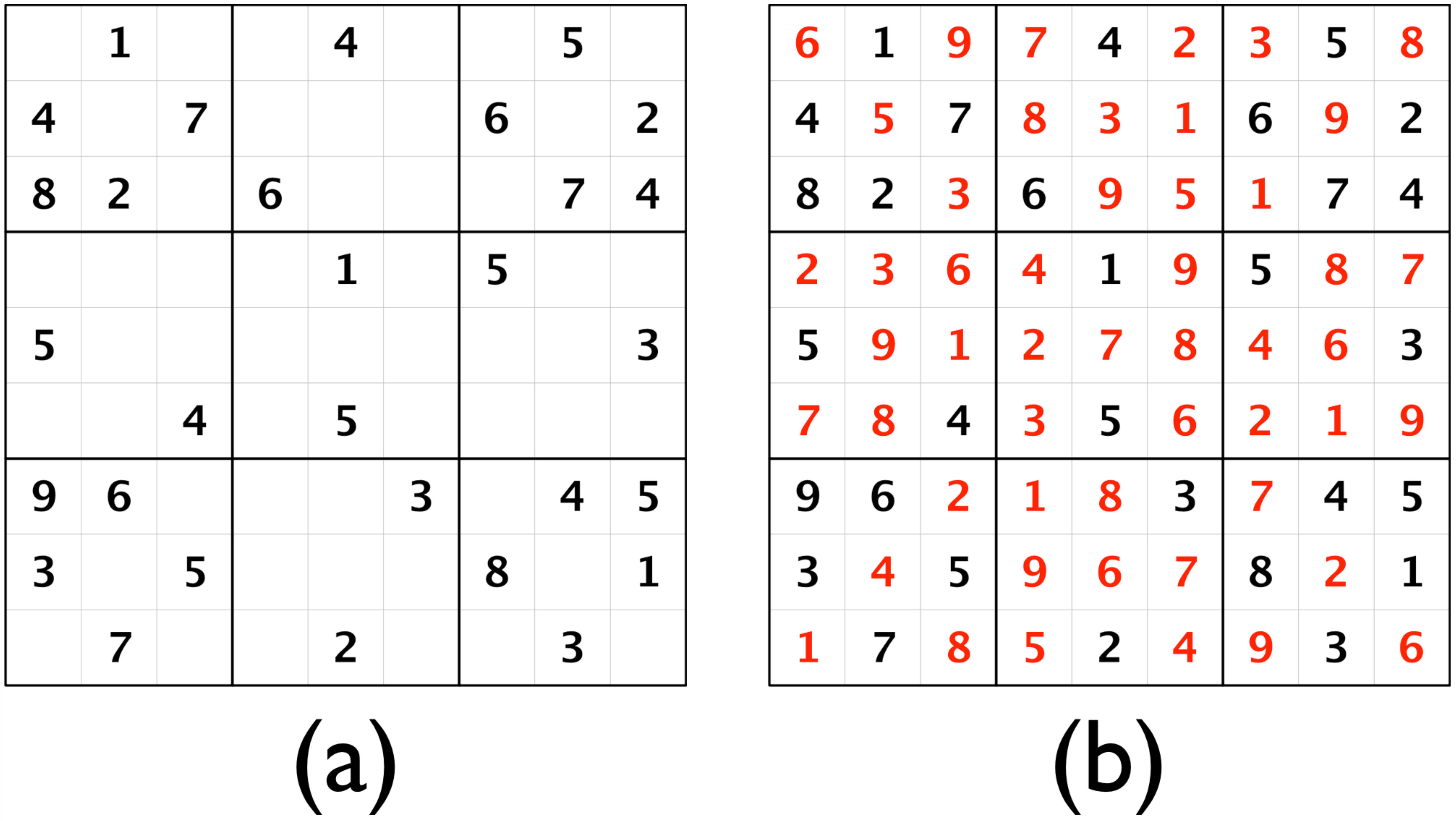}
\end{center}
\caption{A typical ${9}\times{9}$ sudoku puzzle: (a) original problem and (b) corresponding solution, with added digits in red.}
\label{fig:sudoku}
\end{figure}

\subsection{Sudoku}

A Sudoku puzzle is a partially completed row-column grid of cells partitioned into $N$ regions, each of size $N$ cells, to be filled in using a prescribed set of $N$ distinct symbols, such that each row, column, and region contains exactly one of each element of the set.
A well-formed Sudoku puzzle has exactly one solution.
Sudoku is an example of an exact-cover constraint-satisfaction problem and is NP-complete when generalized to ${N}\times{N}$ grids \cite{Yato2003}.

People typically solve Sudoku puzzles on a ${9}\times{9}$ grid (e.g. see Figure \ref{fig:sudoku}) containing nine ${3}\times{3}$ regions, but larger square-in-square puzzles are also possible.
To represent an ${N}\times{N}$ square-in-square Sudoku puzzle as an optimization problem we use $\mathcal{O}(N^3)$ \emph{binary indicator} variables (each serving as a boolean flag) and $\mathcal{O}(N^2)$ hard constraints.
For all open cells (those that have not been supplied as ``clues''), we use a binary indicator variable, designated as $v({row}, {column}, {digit})$, to represent each possible digit assignment. 
For example, the variables $v(1, 3, 1)$, $v(1, 3, 2)$ \ldots, $v(1, 3, 9)$ represent that the cell in row 1, column 3 can take values 1 through 9.

For factor nodes, we developed hard ``one-on'' constraints: a \emph{one-on} constraint requires that a single variable is ``on'' (value = 1.0) and any remaining are ``off'' (value = 0.0).
Furthermore, when appropriate, these constraints output weights that reflect logical certainty: if there is only a single viable output (e.g. all variables but one are known to be ``off''), the outgoing weights are set to $\infty$ (otherwise, 1.0).
We apply \emph{one-on} constraints to four classes of variable sets: one digit assignment per cell; one of each digit assigned per row; one of each digit assigned per column; and one of each digit assigned per square.
Prior work on formulating Sudoku puzzles as constraint-satisfaction problems (e.g. \citeauthor{Simonis2005} \citeyear{Simonis2005}) has utilized additional, redundant constraints to strengthen deduction by combining several of the original constraints, but we only utilize this base constraint set, such as to focus on the effects of reasoners and graph dynamics.

Sudoku is a non-convex problem, so the TWA is not guaranteed to solve the problem.
However, prior work has shown that it is effective for many puzzles, even those that are very large, and that incorporating certainty weights leads to a certainty-propagation algorithm falling out of the TWA \cite{Derbinsky2013AnImp}.
The TWA with certainty weights proceeds in two phases: first, $\infty$ weights from initial clues quickly propagate through the one-on constraints, and then any remaining cells that could not be logically deduced are searched numerically via optimization message-passing.
We now focus on how knowledge-integration can improve performance of the second phase for very difficult puzzles.

\subsubsection{Integrating Knowledge}

We added to the problem graph one local reasoner per cell with the express purpose of maintaining the set of \emph{possible} digit assignments (i.e. those values for which there was not a \emph{certain} message for the ``off'' binary indicator variable) and, when a possibility was removed, communicate this information to a global reasoner.

The global reasoner utilized this information to improve solver efficiency and performance by utilizing graph dynamics to reduce the problem search space.
Specifically, when a possibility was no longer viable logically, the global reasoner removed four edges (those that connected the binary indicator variable to the constraints of the \{cell, row, column, and square\}), as well as any factors that were left option-less in the process).
These graph dynamics were intended to reduce the problem-graph size by removing constraints, as they became unnecessary, and thereby dramatically improve iteration time of the second phase of numerical solving, which is the major time sink for difficult puzzles.

\begin{figure}[b]
\begin{center}
\includegraphics[width=.99\columnwidth]{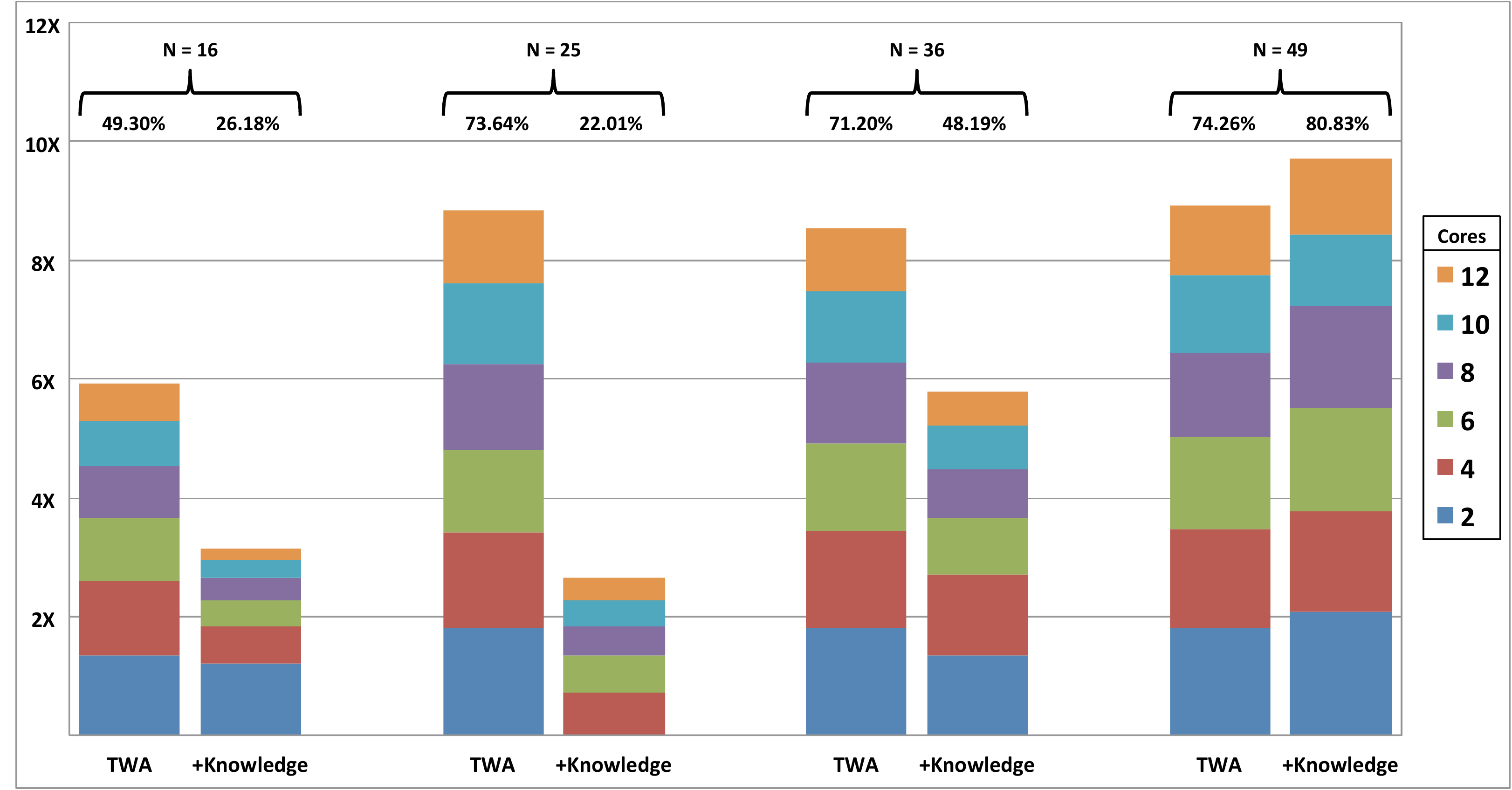}
\end{center}
\caption{Degree of concurrency when solving Sudoku puzzles of different sizes without/with knowledge integration. Improving linearly yields 1 vertical-unit block, and thus ideal concurrency would fill the full vertical space.}
\label{fig:sudoku-cores}
\end{figure}

\begin{table*}[tb]
\caption{$N$x$N$ square-in-square Sudoku puzzles solved without/with the reasoner hierarchy \& graph dynamics. \\} 
\label{fig:sudoku-results}
\begin{tabular}{ | m{1.3cm} | m{1.5cm} | m{1.8cm} | m{1.8cm} | m{1.8cm} | m{1.8cm} | m{1.8cm} | m{1.8cm} | m{0.1cm} }

\centering \cellcolor{black} \textbf{\textcolor{white}{\bm{$N$}}} & \centering \cellcolor{black} \textbf{\textcolor{white}{\# Puzzles}} & \multicolumn{2}{|c|}{\cellcolor{black} \textbf{\textcolor{white}{Avg. Final Graph Size}}} & \multicolumn{2}{|c|}{\cellcolor{black} \textbf{\textcolor{white}{Avg. Iter. Time (msec/iter)}}} & \multicolumn{2}{|c|}{\cellcolor{black} \textbf{\textcolor{white}{Avg. Solve Time (sec)}}} & \\ \cline{1-8}

\cellcolor{black} & \cellcolor{black} & \centering \footnotesize \textbf{TWA} & \centering \footnotesize \textbf{+Knowledge} & \centering \footnotesize \textbf{TWA} & \centering \footnotesize \textbf{+Knowledge} & \centering \footnotesize \textbf{TWA} & \centering \footnotesize \textbf{+Knowledge} & \\ \cline{1-8}

\centering 16 & \centering 10 & \centering \textcolor{white}{00}5,120 & \centering \textcolor{white}{00,}460 & \centering \textcolor{white}{0}1.48 & \centering \textcolor{white}{0}0.80 & \centering \textcolor{white}{00}0.17 & \centering \textcolor{white}{00}0.12 & \\ \cline{1-8}
\centering 25 & \centering \textcolor{white}{0}9 & \centering \textcolor{white}{0}18,125 & \centering \textcolor{white}{0}3,121 & \centering \textcolor{white}{0}6.19 & \centering \textcolor{white}{0}0.92 & \centering \textcolor{white}{0}75.96 & \centering \textcolor{white}{00}5.25 & \\ \cline{1-8}
\centering 36 & \centering 10 & \centering \textcolor{white}{0}51,840 & \centering \textcolor{white}{0}6,032 & \centering 22.19 & \centering \textcolor{white}{0}4.12 & \centering \textcolor{white}{0}61.36 & \centering \textcolor{white}{00}9.03 & \\ \cline{1-8}
\centering 49 & \centering \textcolor{white}{0}3 & \centering \textcolor{white}{}127,253 & \centering 11,173 & \centering 52.48 & \centering \textcolor{white}{}10.54 & \centering 124.59 & \centering \textcolor{white}{0}20.08 & \\ \cline{1-8}

\end{tabular}
\end{table*}

To evaluate this benefit, we downloaded 32 of the hardest puzzles from an online puzzle repository\footnote{\url{http://www.menneske.no/sudoku/eng}} and Table \ref{fig:sudoku-results} summarizes final problem graph size (factors + variables), iteration-loop time (in msec. per iteration), and total solution time (in seconds) with and without this global reasoner (results were gathered using a Java implementation of the TWA running on a single core of a 3.4GHz i7 CPU with OS X 10.8.3).
We see from this data that as the problem increases in size, the global reasoner maintains a much smaller problem graph and corresponding iteration time: even as the baseline TWA crosses 50 msec. per iteration, a commonly accepted threshold for reactivity in the cognitive-architecture community \cite{Rosenbloom2012,Derbinsky2009Effic,Derbinsky2010Towar}, our reasoner is able to easily maintain a real-time response rate for very large sudoku puzzles.
Because our global reasoner modifies the problem graph, and thus the numeric optimizer must now search over a different problem, it was not obvious whether there would be an overall benefit in terms of time-to-solution.
However, the last two columns show that adding knowledge always resulted in a faster solve - up to more than 14$\times$ faster for $N=25$.

We also evaluated a concurrent implementation of the TWA that divides factors/variables into work queues using a na\"{i}ve scheduling algorithm: cost $\propto$ number of incoming edges and we do \emph{not} reschedule work (i.e. queues can become unbalanced).
Figure \ref{fig:sudoku-cores} illustrates the degree to which this implementation can exploit added cores: each stacked segment represents the proportion of speedup in iteration-loop time (msec/iter), as calculated by $( ( \text{time with 1 core} / \text{time with n cores} ) / \text{n} )$, and each vertical tick is 1 unit tall (thus perfect linear speedup would be represented by a stack that spans the full vertical space).
This data reveals two trends: (1) adding cores yields better performance; and (2) our implementation of the TWA can take greater advantage of parallelism with larger problem graphs.
The only exception to the former point was for $N=25, {cores}=2$, in which the overhead of threading actually hurt performance (shown as 0 height in the figure).
The latter point is especially relevant for evaluating graph dynamics, as Figure \ref{fig:sudoku-cores} demonstrates significantly reduced concurrent bandwidth when the problem graph is kept very small ($N \in \{16, 25\}$): however, when comparing the best average time-per-iteration, using graph dynamics was $6.6\times$ faster for $N=49$, $4.5\times$ faster for $N=36$,  $3.5\times$ faster for $N=25$, and was nearly unchanged for $N=16$.

Thus, to summarize our work in Sudoku:

\begin{itemize}
\item{\textbf{Local reasoners} were used to create an \emph{efficient} and \emph{scalable} discrimination network of possibility-set changes for consumption by a global reasoner.}
\item{A \textbf{global reasoner} responded to \emph{changes} in possibility-sets in order to implement graph dynamics.}
\item{These \textbf{graph dynamics} dramatically pruned the problem set size (more than an order of magnitude for large puzzles) and had the result of improving \emph{efficiency} and \emph{scaling}, including reactive iterations for the largest puzzles.}
\end{itemize}

\subsection{Circle Packing}

\begin{figure}[b]
\begin{center}
\includegraphics[width=.6\columnwidth]{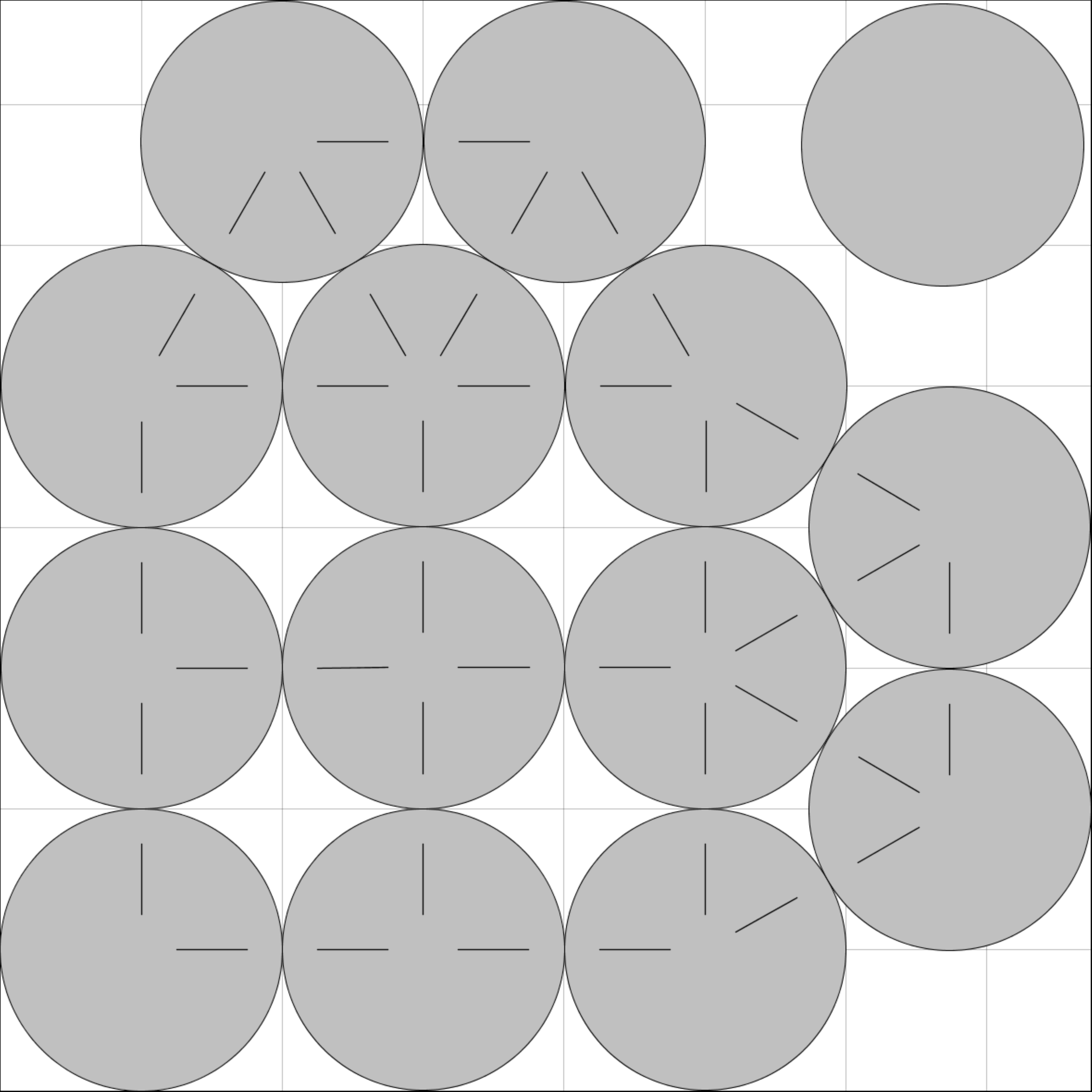}
\end{center}
\caption{An \emph{optimal} packing of 14 circles within a square. Contacting circles are indicated by lines.}
\label{fig:packing-example}
\end{figure}

Circle packing is the problem of positioning a given number of congruent circles in such a way that the circles fit fully in a square without overlapping.
A large number of circles makes finding a solution difficult, due in part to the coexistence of many different circle arrangements with similar density.
For example, Figure \ref{fig:packing-example} shows an optimal packing for 14 circles, which can be rotated across either or both axes, and the free circle in the upper-right corner (a ``rattle'') can be moved without affecting the density of the configuration.

To represent a circle-packing instance with $N$ objects as an optimization problem we use $\mathcal{O}(N)$ continuous variables and $\mathcal{O}(N^2)$ constraints.
Each object has 2 variables: one representing each of its coordinates (or, more generally, $d$ variables for packing spheres in $d$ dimensions).
For each object we create a single \emph{box-intersection} constraint, which enforces that the object stays within the box.
Furthermore, for each pair of objects, we create a \emph{pairwise-intersection} constraint, which enforces that no two objects overlap.
For both sets of constraints, we utilize zero-weight messages when the constraint is not violated (i.e. the object is within the box and/or the two objects are not intersecting).

Like Sudoku, circle packing is a non-convex problem, and thus the TWA is not guaranteed to converge.
However, prior work has shown that the TWA is effective and that using zero-weight messages to ``ignore'' messages from inactive constraints dramatically reduces interference and thus the number of iterations to convergence \cite{Derbinsky2013AnImp}.

\subsubsection{Integrating Knowledge}

We made use of our knowledge-integration methods for two distinct purposes: (1) drastically improved scaling via r-tree integration and (2) macro movements via human assistance.

Using zero-weight messages for inactive constraints drastically improves iterations-to-convergence for circle packing; however, this method still requires $\mathcal{O}(N^2)$ intersection constraints, and thus iteration-loop time (msec/iteration) still prevents scaling to large numbers of circles.
However, there are numerous methods for efficiently detecting intersections of spatial objects, such as r-trees \cite{Guttman1984}, and since the TWA tends to move circles in a local fashion, we implemented a graph-dynamics global reasoner that integrates an r-tree to add/remove intersection constraints each iteration (see Algorithm \ref{alg:cp}).
To begin, we update the r-tree with the locations of all circles via concurred variable values (line 1).
We use axis-aligned bounding boxes (AABBs) to describe object locations and add a 5\% ``neighborhood'' buffer to each box: empirically we found that if constraints were immediately removed from neighboring circles, there was a high probability of cycles and thrashing, but by keeping a small constant set of adjoining circles, the problem-graph size was reasonably small (though non-decreasing for reasonable densities) and the solver was robust to starting conditions and configuration densities.
We then query for all object intersections (line 2) and categorize them as either existing (line 4) or new (line 5), based upon the current graph.
We then iterate over factors in the problem graph (line 9): for existing intersections with corresponding factors, we do nothing (line 11); and for those factors that are no longer intersecting (or in the neighborhood buffer), we remove those factors from the graph and add to a ``pool'' of constraints (lines 13-14).
Finally, for all new intersections, we add a constraint to the graph, drawing from, and re-configuring, past [and now unused] factors, if available in the pool (lines 18-19).
The key insights of this algorithm are as follows: (1) only those objects that are in the graph need to be updated in the r-tree, as those are the only coordinates the TWA could have altered; (2) the algorithm scales with the number of \emph{active} factors and intersections, which should be small compared to the full $\mathcal{O}(N^2)$ set\footnote{With a relatively small neighborhood buffer, the size of the active set is $\mathcal{O}(N)$ and related to the \emph{kissing number}.}; and (3) we pool and re-parameterize unused factors to bound memory consumption.

\begin{algorithm}
UpdateRTree()\;
\ForEach{intersection}{
\eIf{in graph}{Touch()\;}{AddToAdditionQueue()\;}
}
\ForEach{intersection factor}{
\eIf{touched}{do nothing\;}{
RemoveFromGraph()\;
AddToFactorPool()\;
}
}
\ForEach{factor in addition queue}{
GetFactorFromPool()\;
AddToGraph()\;
}
\caption{Core logic for a global reasoner implementing dynamic-graph maintenance in the circle-packing task.}
\label{alg:cp}
\end{algorithm}

Using this approach, we have scaled our circle-packing implementation to more than 2 million circles, which is two orders of magnitude more than previously reported results\footnote{http://www.packomania.com}.
We have also achieved a record-breaking density of 0.82 with 9,996 circles in a unit square, the largest previously reported problem instance (previous density $< 0.81$).
For that problem, average iteration-loop time was less than 80 msec./iteration (4 cores, 3.4GHz i7) and the problem graph never exceeded 1GB of memory, representing less than 0.09\% of the $\sim{50M}$ nodes in the full problem.

In watching the iteration dynamics of circle packing, we noticed that because the TWA makes local changes to circle positions, it can be difficult for the algorithm to make larger, coordinated movements to escape local minima.
Thus, we implemented two sets of reasoners to incorporate human assistance, via a mouse, into the low-level circle-packing optimization.
The first implementation required only a single local reasoner: when a user clicked and dragged a circle in the GUI (which reflected iteration results in real-time), the mouse coordinates were sent as weighted messages to that circle's coordinate variables (graph dynamics were used to connect the local-reasoner factor on-demand).
Empirically, we found that if human users knew high-level properties of the final packing (e.g. they had seen a picture of a final configuration), they could easily communicate the macro structure via a series of circle movements, and then the TWA would integrate this information and perform low-level, fine-grained movements.
The result was that humans working with the TWA could consistently achieve packings that were near record-breaking, and the human users would often interpret the experience as ``fun'' and ``game-like.''

This type of interaction, however, did not work well when we increased the number of circles to many thousands or millions of circles.
However, we still found that humans could contribute high-level knowledge that was very useful, but difficult to compute algorithmically within the TWA: areas with free space.
Thus, we implemented a reasoner hierarchy: a global reasoner extracted meta-data from the r-tree intersections to identify the circle with the \emph{largest} current overlap, the user would click a point in the GUI to identify a region of free space, and a local reasoner would \emph{transport} (via weighted messages) the ``distressed'' circle to the ``vacant'' region.
Empirically this implementation seemed to decrease iterations-to-convergence, but we have not verified this hypothesis in a controlled setting.

For both of these reasoner implementations, we asked humans to contribute high-level task knowledge in the form of mouse clicks.
However, it is conceivable that human perceptual and motor models, such as those in EPIC \cite{Kieras1997Overview}, might achieve similar results in an automated fashion and we are interested to further study this possibility.

Thus, to summarize our work in circle packing:

\begin{itemize}
\item{A \textbf{local reasoner} \emph{integrated} mouse input from humans.}
\item{A \textbf{global reasoner} integrated an r-tree with the main iteration loop for greatly improved \emph{efficiency} and problem-size \emph{scaling}, as well as to inform human-assisted packing via \emph{integration} of intersection meta-data.}
\item{\textbf{Graph dynamics} maintained the set of constraints, as informed by an r-tree. These constraints were re-parameterized to \emph{bound memory usage}.}
\end{itemize}


\section{Discussion}

The focus of this paper was to consider whether general optimization could serve as a useful platform upon which to quickly and flexibly develop a variety of cognitive-processing modules.
In this context, we presented the Three-Weight Algorithm as a candidate approach, along with novel methods by which to usefully interface high-level knowledge with a low-level optimization framework in order to improve expressiveness of knowledge, methods, and heuristics, as well as bolster algorithm efficiency scaling.
In order to exemplify these methods, we employed two tasks, independent of an agent or cognitive architecture.

Future work needs to proceed down [at least] three separate paths.
First, these methods need to be evaluated within actual cognitive-processing modules.
For example, cognitive modelers could benefit from memory modules that are flexible (via an arbitrary objective function) but are also efficient for real-time use and scalable to complex tasks, which might be accomplished via global reasoners that exploit state-of-the-art indexing techniques \cite{Derbinsky2010Towar}.
Furthermore, these types of methods seem to lend themselves to exploring the interface between state-of-the-art perception algorithms and a symbolic cognitive architecture.
Second, the TWA and our knowledge-integration techniques need to be evaluated in context of a running agent, with real-time environmental input and changing knowledge, goals, and preferences.
Finally, we need to explore whether effective and efficient learning can occur within modules that employ these methods.


\bibliographystyle{aaai}
\bibliography{ref}

\end{document}